\documentclass{article}
\PassOptionsToPackage{numbers, compress}{natbib}

    \usepackage[preprint]{neurips_2020}

\usepackage[utf8]{inputenc} 
\usepackage[T1]{fontenc}    
\usepackage{hyperref}       
\usepackage{url}            
\usepackage{booktabs}       
\usepackage{amsfonts}       
\usepackage{nicefrac}       
\usepackage{microtype}      

\usepackage{mathrsfs} 
\usepackage{amsmath} 

\usepackage{enumerate}

\usepackage{booktabs}
\usepackage{multirow}
\usepackage{floatrow}

\newfloatcommand{capbtabbox}{table}[][\FBwidth]

\usepackage{algorithmic}
\usepackage{algorithm}

\usepackage[pdftex]{graphicx}
\usepackage{subfigure}

\title{News-Driven Stock Prediction With Attention-Based Noisy Recurrent State Transition}

\author{
	Xiao Liu\textsuperscript{1}, Heyan Huang\textsuperscript{1}, Yue Zhang\textsuperscript{2}\thanks{Corresponding author.}, Changsen Yuan\textsuperscript{1}\\
  \textsuperscript{1}School of Computer Science and Technology, Beijing Institute of Technology \\
  \textsuperscript{2}School of Engineering, Westlake University \\
  \textsuperscript{3}Institute of Advanced Technology, Westlake Institute for Advanced Study \\
  {\tt \{xiaoliu,hhy63,yuanchangsen\}@bit.edu.cn} \\
  {\tt yue.zhang@wias.org.cn}
}

\begin{document}

\maketitle

\begin{abstract}
  We consider direct modeling of underlying stock value movement sequences over time in the news-driven stock movement prediction.
  A recurrent state transition model is constructed, which better captures a gradual process of stock movement continuously by modeling the correlation between past and future price movements.
  By separating the effects of news and noise, a noisy random factor is also explicitly fitted based on the recurrent states.
  Results show that the proposed model outperforms strong baselines.
  Thanks to the use of attention over news events, our model is also more explainable.
  To our knowledge, we are the first to explicitly model both events and noise over a fundamental stock value state for news-driven stock movement prediction.
\end{abstract}

\section{Introduction}
Stock movement prediction is a central task in computational and quantitative finance.
With recent advances in deep learning and natural language processing technology, event-driven stock prediction has received increasing research attention \citep{DBLP:conf/acl/XiePWC13,DBLP:conf/ijcai/DingZLD15}.
The goal is to predict the movement of stock prices according to financial news.
Existing work has investigated news representation using bag-of-words \citep{DBLP:conf/naacl/KoganLRSS09}, named entities \citep{DBLP:journals/tois/SchumakerC09}, event structures \citep{DBLP:conf/emnlp/DingZLD14} or deep learning \citep{DBLP:conf/ijcai/DingZLD15,DBLP:conf/acl/CohenX18}.

Most previous work focuses on enhancing news representations, while adopting a relatively simple model on the stock movement process, casting it as a simple response to a set of historical news.
The prediction model can therefore be viewed as variations of a classifier that takes news as input and yields stock movement predictions.
In contrast, work on time-series based stock prediction \citep{10.1093/wber/10.2.323,AMIHUD200231,DBLP:conf/acl/CohenX18,DBLP:journals/corr/abs-1809-00306}, aims to capture continuous movements of prices themselves.

We aim to introduce underlying price movement trends into news-driven stock movement prediction by casting the underlaying stock value as a recurrent state, integrating the influence of news events and random noise simultaneously into the recurrent state transitions.
In particular, we take a LSTM with peephole connections \citep{DBLP:conf/ijcnn/GersS00} for modeling a stock value state over time, which can reflect the fundamentals of a stock.
The influence of news over a time window is captured in each recurrent state transition by using neural attention to aggregate representations of individual news.
In addition, all other factors to the stock price are modeled using a random factor component, so that sentiments, expectations and noise can be dealt with explicitly.

\begin{figure}
\centering
\includegraphics[width=7.3cm]{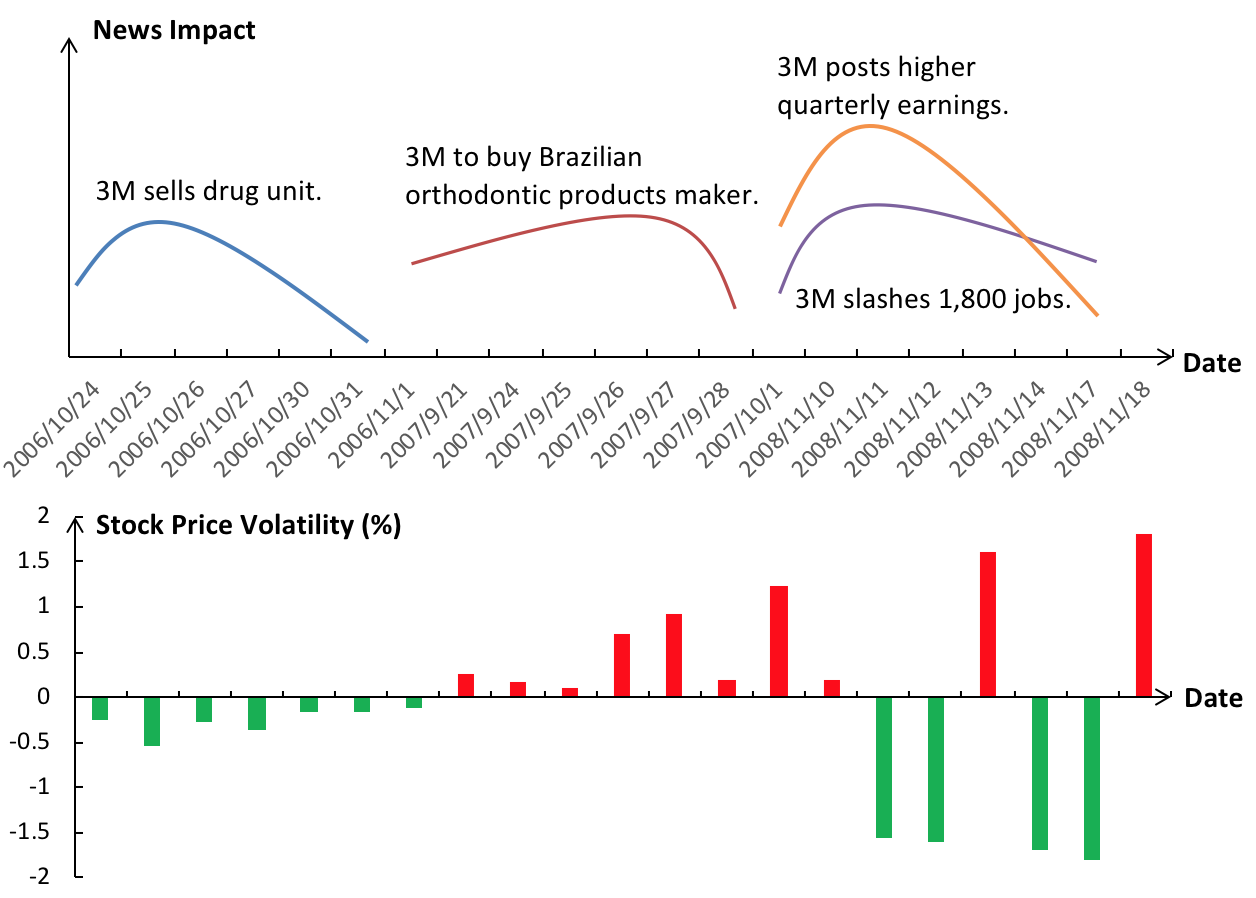}
\vspace{-0.1cm}
\caption{
\label{fig:impact}
Example of news impacts on \textit{3M Company}.
Over the first and the second periods (from Oct. 24 to Nov. 1, 2006 and from Sep. 21 to Oct. 1, 2007), there was only one event.
In the third period (from Nov. 10 to Nov. 18, 2008), there were two events affecting the stock price movements simultaneously.
}
\end{figure}

Compared with existing work, our method has three salient advantages.
First, the process in which the influence of news events are absorbed into stock price changes is explicitly modeled.
Though previous work has attempted towards this goal \citep{DBLP:conf/ijcai/DingZLD15}, existing models predict each stock movement independently, only modeling the correlation between news in historical news sequences.
As shown in Figure \ref{fig:impact}, our method can better capture a continuous process of stock movement by modeling the correlation between past and future stock values directly.
In addition, non-linear compositional effects of multiple events in a time window can be captured.

Second, to our knowledge, our method allows noise to be explicitly addressed in a model, therefore separating the effects of news and other factors.
In contrast, existing work trains a stock prediction model by fitting stock movements to events, and therefore can suffer from overfitting due to external factors and noise.

Third, our model is also more explainable thanks to the use of attention over news events, which is similar to the work of \citep{DBLP:conf/coling/ChangZTBK16} and \citep{DBLP:journals/corr/abs-1902-04994}.
Due to the use of recurrent states, we can visualize past events over a large time window.
In addition, we propose a novel future event prediction module to factor in likely next events according to natural events consequences.
The future event module is trained over gold ``future'' data over historical events.
Therefore, it can also deal with insider trading factors to some extent.

Experiments over the benchmark of \citep{DBLP:conf/ijcai/DingZLD15} show that our method outperforms strong baselines, giving the best reported results in the literature.
To our knowledge, we are the first to explicitly model both events and noise over a fundamental stock value state for news-driven stock movement prediction.
Note that unlike time-series stock prediction models \citep{DBLP:journals/kbs/ZhangZWYFY18,DBLP:conf/acl/CohenX18}, we do not take explicit historical prices as part of model inputs, and therefore our research still focuses on the influence of news information alone, and are directly comparable to existing work on news-driven stock prediction.

\section{Related Work}
There has been a line of work predicting stock markets using text information from daily news.
We compare this paper with previous work from the following two perspectives.

\vspace{1mm}
\noindent\textbf{Modeling Price Movements Correlation}

Most existing work treats the modeling of each stock movement independently using bag-of-words \citep{DBLP:conf/naacl/KoganLRSS09}, named entities \citep{DBLP:journals/tois/SchumakerC09}, semantic frames \citep{DBLP:conf/acl/XiePWC13}, event structures \citep{DBLP:conf/emnlp/DingZLD14}, event embeddings \citep{DBLP:conf/ijcai/DingZLD15} or knowledge bases \citep{DBLP:conf/coling/DingZLD16}.
Differently, we study modeling the correlation between past and future stock value movements.

There are also some work modeling the correlations between samples by sparse matrix factorization \citep{DBLP:conf/icdm/WongLC14}, hidden Markov model \citep{DBLP:journals/corr/abs-1809-00306} and Bi-RNNs \citep{DBLP:conf/acl/CohenX18,DBLP:journals/corr/abs-1902-04994} using both news and historical price data.
Some work models the correlations among different stocks by pre-defined correlation graph \citep{DBLP:conf/naacl/PengJ16} and tensor factorization \citep{DBLP:journals/kbs/ZhangZWYFY18}.
Our work is different from this line of work in that we use only news events as inputs, and our recurrent states are combined with impact-related noises.


\vspace{1mm}
\noindent\textbf{Explainable Prediction}

Rationalization is an important problem for news-driven stock price movement prediction, which is to find the most important news event along with the model's prediction.
Factorization, such as sparse matrix factorization \citep{DBLP:conf/icdm/WongLC14} and tensor factorization \citep{DBLP:journals/kbs/ZhangZWYFY18}, is a popular method where results can be traced back upon the input features.
While this type of method are limited because of the dimension of input feature, our attention-based module has linear time complexity on feature size.

\citep{DBLP:journals/corr/abs-1902-04994} apply dual-layer attention to predict the stock movement by using news published in the previous six days.
Each day's news embeddings and seven days' embeddings are summed by the layer.
Our work is different from \citep{DBLP:journals/corr/abs-1902-04994} in that our news events attention is query-based, which is more strongly related to the noisy recurrent states.
In contrast, their attention is not query-based and tends to output the same result for each day even if the previous day's decision is changed.

\begin{figure*}
\centering
\includegraphics[width=12.5cm]{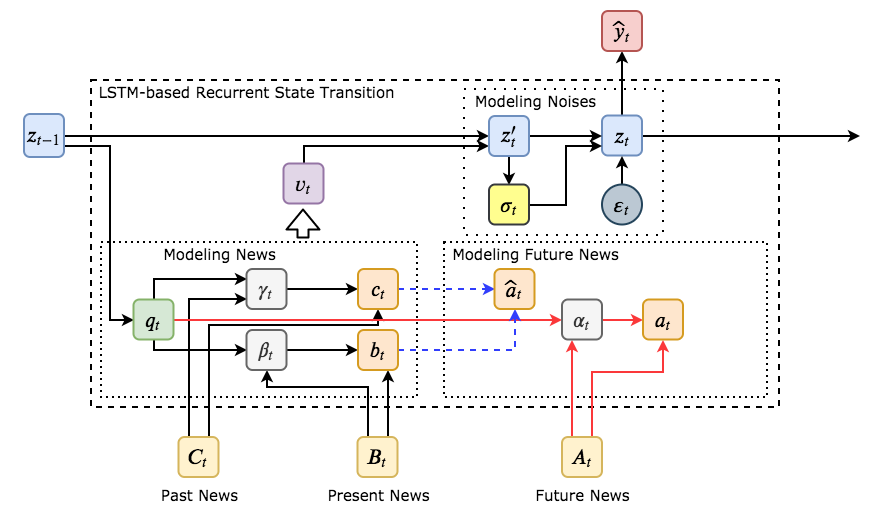}
\vspace{-0.1cm}
\caption{
The ANRES model framework for trading day $t$ in a trading sequence.
The black solid elbows are used both in the training and the evaluating procedures.
The red solid elbows are only used in the training procedure, while the blue dotted elbows in the evaluating procedure.
}
\label{fig:model}
\end{figure*}

\section{Task Definition}
Following previous work \citep{DBLP:conf/emnlp/DingZLD14,DBLP:conf/ijcai/DingZLD15}, the task is formalized as a binary classification task for each trading day.
Formally, given a history news set about a targeted stock or index, the input of the task is a trading day $x$ and the output is a label $y \in \{+1, -1\}$ indicating whether the adjusted closing price $p_x$ will be greater than $p_{x-1}$ ($y=+1$) or not ($y=-1$).

\section{Method}
The framework of our model is shown in Figure \ref{fig:model}.
We explicitly model both events and noise over a recurrent stock value state, which is modeled using LSTM.
For each trading day, we consider the news events happened in that day as well as the past news events using neural attention \citep{DBLP:conf/nips/VaswaniSPUJGKP17}.
Considering the impacts of insider trading, we also involve future news in the training procedure.
To model the high stochasticity of stock market, we sample an additive noise using a neural module.
Our model is named attention-based noisy recurrent states transition (ANRES).

Considering the general principle of sample independence, building temporal connections between individual trading days in training set is not suitable for training \citep{DBLP:conf/acl/CohenX18} and we find it easy to overfit.
We notice that a LSTM usually takes several steps to generate a more stable hidden state.
As an alternative method, we extended the time span of one sample to $T$ previous continuous trading days (${t-T+1, t-T+2, ..., t-1, t}$), which we call a trading sequence, is used as the basic training element in this paper.

\subsection{LSTM-based Recurrent State Transition}
ANRES uses LSTM with peephole connections \citep{DBLP:conf/ijcnn/GersS00}.
The underlying stock value trends are represented as a recurrent state $z$ transited over time, which can reflect the fundamentals of a stock.
In each trading day, we consider the impact of corresponding news events and a random noise as:
\begin{align*}
	z^{\prime}_t &= \overrightarrow{\textrm{LSTM}}(v_t, z_{t-1})  \\
	z_t &= f(z^{\prime}_t)           
\end{align*}
where $v_t$ is the news events impact vector on the trading day $t$ and $f$ is a function in which random noise will be integrated.

By using this basic framework, the non-linear compositional effects of multiple events can also be captured in a time window.
Then we use the sequential state $z_t$ to make binary classification as:
\begin{align*}
	\hat{p}_t &= \textrm{softmax}(W^y z_t)  \\
	\hat{y}_t &= \mathop{\arg\max}^{i}_{i \in \{+1, -1\}} \hat{p}_t(\hat{y}_t=i|x_t) 
\end{align*}
where $\hat{p}_t$ is the estimated probabilities, $\hat{y}_t$ is the predicted label and $x_t$ is the input trading day.

\subsection{Modeling News Events}
For a trading day $t$ in a trading sequence, we model both long-term and short-term impact of news events.
For short-term impact, we use the news published after the previous trading day $t-1$ and before the trading day $t$ as the present news set.
Similarly, for long-term impact, we use the news published no more than thirty calendar days ago as the past news set.

For each news event, we extract its headline and use ELMo \citep{DBLP:conf/naacl/PetersNIGCLZ18} to transform it to $V$-dim hidden state by concatenating the output bidirectional hidden states of the last words as the basic representation of a news event.
By stacking those vectors accordingly, we obtain two embedding matrices $C^{\prime}_t$ and $B^{\prime}_t$ for the present and past news events as:
\begin{align*}
{ec}^i_t &= \overleftrightarrow{\textrm{ELMo}}({hc}^i_t), i \in \{1, 2, ..., L_c\} \\
{eb}^j_t &= \overleftrightarrow{\textrm{ELMo}}({hb}^j_t), j \in \{1, 2, ..., L_b\} \\
C^{\prime}_t &= \textrm{stack}(\{{ec}^1_t, {ec}^2_t, ..., {ec}^{L_c}_t\}) \\
B^{\prime}_t &= \textrm{stack}(\{{eb}^1_t, {eb}^2_t, ..., {eb}^{L_b}_t\})
\end{align*}
where ${hc}^i_t$ is one of the news event headline in the present news set, ${ec}^i_t$ is the headline representation of ${hc}^i_t$, $L_c$ is the size of present news set; while ${hb}^j_t$, ${eb}^j_t$ and $L_b$ are for the past news set.

To make the model more numerically stable and avoiding overfitting, we apply the over-parameterized component of \citep{DBLP:journals/corr/abs-1911-11423} to the news events embedding matrices, where
\begin{align*}
C_t &= \sigma(W^f C^{\prime}_t) \odot \textrm{tanh}(W^c C^{\prime}_t) \\
B_t &= \sigma(W^f B^{\prime}_t) \odot \textrm{tanh}(W^c B^{\prime}_t)
\end{align*}

$\odot$ is element-wise multiplication and $\sigma(\cdot)$ is the sigmoid function.

Due to the unequal importance news events contribute to the stock price movement in $t$, we use scaled dot-product attention \citep{DBLP:conf/nips/VaswaniSPUJGKP17} to capture the influence of news over a period for the recurrent state transition.
In practical, we first transform the last trading day's stock value $z_{t-1}$ to a query vector $q_t$, and then calculate two attention score vectors $\gamma_t$ and $\beta_t$ for the present and past news events as:
\begin{align*}
q_t &= \textrm{tanh}(W^q z_{t-1}) \\
\gamma_t &= \textrm{softmax}(\frac{C_t^i q_t}{\sqrt{V}}) \\
\beta_t &= \textrm{softmax}(\frac{B_t^i q_t}{\sqrt{V}})
\end{align*}

We sum the news events embedding matrices to obtain news events impact vectors $c_t$ and $b_t$ on the trading day $t$ according to the weights $\gamma_t$ and $\beta_t$, respectively:
\begin{align*}
c_t &= \textrm{tanh}(\sum^{N_t}_{i=1} \gamma_t^i C_t^i) \\
b_t &= \textrm{tanh}(\sum^{N_t}_{i=1} \beta_t^i B_t^i)
\end{align*}

\subsection{Modeling Future News}
In spite of the long-term and short-term impact, we find that some short-term future news events will exert an influence on the stock price movement before the news release, which can be attributed to news delay or insider trading \citep{10.2307/2118390} factors to some extent.

We propose a novel future event prediction module to consider likely next events according to natural consequences.
In this paper, we define future news events as those that are published within seven calendar days after the trading day $t$.

Similarly to the past and present news events, we stack the headline ELMo embeddings of future news events to an embedding matrix $A^{\prime}_t$.
Then adapting the over-parameterized component and summing the stacked embedding vectors by scaled dot-product attention.
We calculate the future news events impact vector $a_t$ on the trading day $t$ as:
\begin{align*}
A_t &= \sigma(W^f A^{\prime}_t) \cdot \textrm{tanh}(W^c A^{\prime}_t) \\
\alpha_t &= \textrm{softmax}(\frac{A_t^i q_t}{\sqrt{V}}) \\
a_t &= \textrm{tanh}(\sum^{N_t}_{i=1} \alpha_t^i A_t^i)
\end{align*}

Although the above steps can work in the training procedure, where the future event module is trained over gold ``future'' data over historical events, at test time, future news events are not accessible.
To address this issue, we use a non-linear transformation to estimate a future news events impact vector $\hat{a}_t$ with the past and present news events impact vectors $b_t$ and $c_t$ as:
\begin{align*}
\hat{a}_t = \textrm{tanh}(W^a [c_t, b_t])
\end{align*}
where $[,]$ is the vector concatenation operation.

We concatenate the above-mentioned three types of news events impact vectors to obtain the input $v_t$ for LSTM-based recurrent state transition on trading day $t$ as:
\begin{align*}
v_t=\left\{
\begin{array}{rcl}
{[c_t, b_t, a_t]}, & & \textrm{when \ training}\\
{[c_t, b_t, \hat{a}_t]}, & & \textrm{when \ evaluating}
\end{array} \right.
\end{align*}
where $[,]$ is the vector concatenation operation.

\subsection{Modeling Noise}
In this model, all other factors to the stock price such as sentiments, expectations and noise are explicitly modeled as noise using a random factor.
We sample a random factor from a normal distribution $\mathcal N(\textbf{0}, \sigma_t)$ parameterized by $z^{\prime}_t$ as:
\begin{align*}
\sigma_t &= \sqrt{\textrm{exp}(\textrm{tanh}(W^{\sigma} z^{\prime}_t))}
\end{align*}

However, in practice, the model can face difficulty of back propagating gradients if we directly sample a random factor from $\mathcal N(\textbf{0}, \sigma_t)$.
We use re-parameterization \citep{DBLP:conf/iclr/SrivastavaS17} for normal distributions to address the problem and enhance the transition result $z^{\prime}_t$ with sample random factor to obtain the noisy recurrent state $z_t$ as:
\begin{align*}
\epsilon_t &\sim \mathcal N(\textbf{0}, \textbf{1}) \\
z_t &= \textrm{tanh}(z^{\prime}_t + \sigma_t \epsilon_t)	
\end{align*}

\subsection{Training Objective}
For training, there are two main terms in our loss function.
The first term is a cross entropy loss for the predicted probabilities $\hat{p}_t$ and gold labels $y_t$, and the second term is the mean squared error between the estimated future impact vector $\hat{a}_t$ and the true future impact vector $a_t$.

The total loss for a trading sequence containing $T$ trading days with standard $L_2$ regularization is calculated as:
\begin{align*}
&L_{ce} = \sum^{T}_{t=1} -\textrm{log}(1 - \hat{p}_t(y_t|x_t)) \\
&L_{mse} = \frac{1}{V} \sum^{T}_{t=1} \sum_{i=1}^{V} (\hat{a}_t^i - a_t^i)^2 \\
&L_{total} = L_{ce} + \theta L_{mse} + \lambda {\| \Phi \|}^2_2
\end{align*}
where $\theta$ is a hyper-parameter which indicates how much important $L_{mse}$ is comparing to $L_{ce}$, $\Phi$ is the set of trainable parameters in the entire ANRES model and $\lambda$ is the regularization weight.

\section{Experiments}

\begin{table}
\centering
\small
\setlength{\tabcolsep}{1mm}{
\begin{tabular}{l|c|c|c}
\hline
 & \textbf{Training} & \textbf{Development} & \textbf{Test} \\
\hline
\#documents & 358,122 & 96,299 & 99,030 \\
\#samples & 1,425 & 169 & 191 \\
time span & 10/20/2006- & 06/19/2012- & 02/22/2013- \\
          & 06/18/2012  & 02/21/2013  & 11/21/2013  \\
\hline
\end{tabular}
}
\vspace{-0.1cm}
\caption{
\label{tab:data_description}
Statistics of the datasets.
}
\end{table}

We use the public financial news dataset released by \citep{DBLP:conf/emnlp/DingZLD14}, which is crawled from Reuters and Bloomberg over the period from October 2006 to November 2013.
We conduct our experiments on predicting the Standard \& Poor’s 500 stock (S\&P 500) index and its selected individual stocks, obtaining indices and prices from Yahoo Finance\footnote{\url{https://finance.yahoo.com/}}.
Detailed statistics of the training, development and test sets are shown in Table \ref{tab:data_description}.
We report the final results on test set after using development set to tune some hyper-parameters.


\subsection{Settings}
The hyper-parameters of our ANRES model are shown in Table \ref{tab:setting}.
We use mini-batches and stochastic gradient descent (SGD) with momentum to update the parameters.
Most of the hyper-parameters are chosen according to development experiments, while others like dropout rate $r$ and SGD momentum $\mu$ are set according to common values.

\begin{table}
\centering
\small
\begin{tabular}{c|c}
\hline
\textbf{Name} & \textbf{Value} \\
\hline
batch size & 16 \\
learning rate $lr$ & 0.005 \\
SGD momentum $\mu$ & 0.9 \\
dropout rate $r$ & 0.3 \\
MSE loss weight $\theta$ & 0.4 \\
regularization weight $\lambda$ & 0.0005 \\
news embedding dimension $V$ & 256 \\
recurrent state dimension $D$ & 100 \\
trading sequence length $T$ & 7 \\
\hline
\end{tabular}
\vspace{-0.1cm}
\caption{\label{tab:setting} Hyper-parameters setting.}
\end{table}

Following previous work \citep{DBLP:conf/acl/XiePWC13,DBLP:conf/emnlp/DingZLD14,DBLP:conf/acl/CohenX18}, we adopt the standard measure of accuracy and Matthews Correlation Coefficient (MCC) to evaluate S\&P 500 index prediction and selected individual stock prediction.
MCC is applied because it avoids bias due to data skew.
Given the confusion matrix which contains true positive, false positive, true negative and false negative values, MCC is calculated as:
\begin{align*}
\textrm{MCC} = \frac{\textrm{tp} \times \textrm{tn} - \textrm{fp} \times \textrm{fn}}{\sqrt{(\textrm{tp} + \textrm{fp})(\textrm{tp} + \textrm{fn})(\textrm{tn} + \textrm{fp})(\textrm{tn} + \textrm{fn})}}
\end{align*}

\subsection{Initializing Noisy Recurrent States}

\begin{table}
\centering
\small
\begin{tabular}{l|c|c}
\hline
 & \textbf{Accuracy} & \textbf{MCC} \\
\hline
ANRES\_Sing\_R & 62.91\% & 0.3704 \\
ANRES\_Sing\_Z & 63.63\% & 0.3672 \\
ANRES\_Seq\_R & 67.94\% & 0.5141 \\
ANRES\_Seq\_Z & \textbf{68.51\%} & \textbf{0.5392} \\
\hline
\end{tabular}
\vspace{-0.1cm}
\caption{
\label{tab:tuning_states}
Development set results on initializing the noisy recurrent states.
}
\end{table}

As the first set of development experiments, we try different ways to initialize the noisy recurrent states of our ANRES model to find a suitable approach.
For each trading day, we compare the results whether states transitions are modeled or not.
Besides, we also compare the methods of random initialization and zero initialization.
Note that the random initialization method we use here returns a tensor filled with random numbers from the standard normal distribution $\mathcal{N}(0, 1)$.
In summary, the following four baselines are designed:
\begin{itemize}
	\item \textit{ANRES\_Sing\_R}: randomly initializing the states for each single trading day.
	\item \textit{ANRES\_Sing\_Z}: initializing the states as zeros for each single trading day. 
	\item \textit{ANRES\_Seq\_R}: randomly initializing the first states for each trading sequence only.
	\item \textit{ANRES\_Seq\_Z}: initializing the first states as zeros for each trading sequence only. 
\end{itemize}

Development set results on predicting S\&P 500 index are shown in Table \ref{tab:tuning_states}.
We can see that modeling recurrent value sequences performs better than treating each trading day separately, which shows that modeling trading sequences can capture the correlations between trading days and the non-linear compositional effects of multiple events.
From another perspective, the models \textit{ANRES\_Sing\_R} and \textit{ANRES\_Sing\_Z} also represent the strengths of our basic representations of news events in isolation.
Therefore, we can also see that using only the basic news events representations is not sufficient for index prediction, while combining with our states transition module can achieve strong results.

By comparing the results of \textit{ANRES\_Seq\_R} and \textit{ANRES\_Seq\_Z}, we decide to use zero initialization for our ANRES models, including the noisy recurrent states also in the remaining experiments.

\subsection{Study on Trading Sequence Length}

\begin{figure}
\centering
\includegraphics[width=7.4cm]{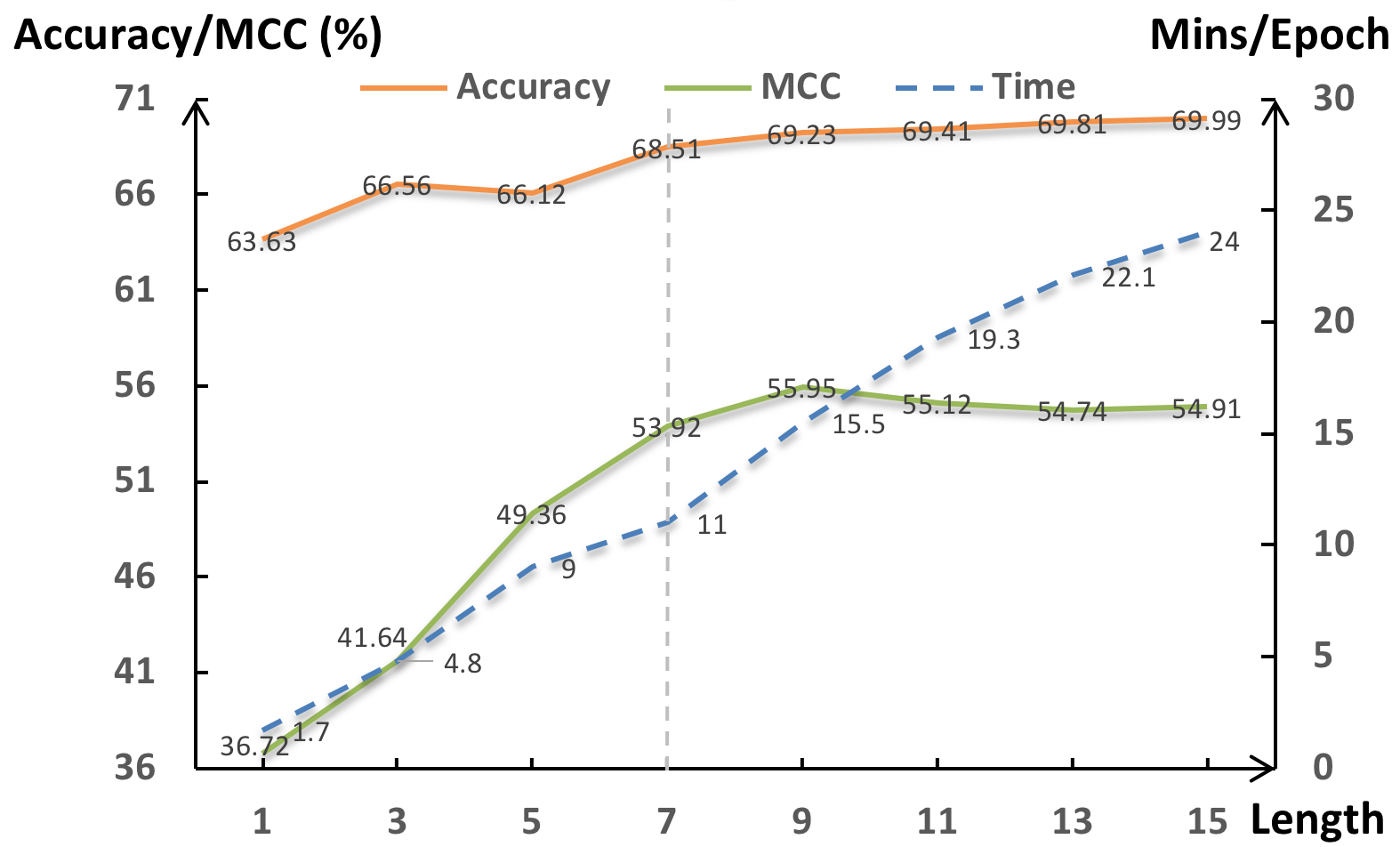}
\vspace{-0.1cm}
\caption{
Development set results of different trading sequence length $T$.
}
\label{fig:tuning_length}
\end{figure}

We use the development set to find a suitable length $T$ for trading sequence, which is searched from $\{1, 3, 5, 7, 9, 11, 13, 15\}$.
The S\&P 500 index prediction results  of accuracy, MCC and consumed minutes per training epoch on the development set are shown in Figure \ref{fig:tuning_length}.

We can see that the accuracy and MCC are positively correlated with the growth of $T$, while the change of accuracy is smaller than MCC.
When $T \geq 7$, the growth of MCC becomes slower than that when $T < 7$.
Also considering the running time per training epoch, which is nearly linear w.r.t. $T$, we choose the hyper-parameter $T=7$ and use it in the remaining experiments.

\subsection{Predicting S\&P 500 Index}
\label{index}

We compare our approach with the following strong baselines on predicting the S\&P 500 index, which also only use financial news:
\begin{itemize}
	\item \citep{luss2015predicting} uses bags-of-words to represent news documents, and constructs the prediction model by using Support Vector Machines (SVMs).
	\item \citep{DBLP:conf/ijcai/DingZLD15} uses event embeddings as input and convolutional neural network prediction model.
	\item \citep{DBLP:conf/coling/DingZLD16} empowers event embeddings with knowledge bases like YAGO and also adopts convolutional neural networks as the basic prediction framework.
	\item \citep{DBLP:conf/acl-alta/PinheiroD17} uses fully connected model and character-level embedding input with LSTM to encode news texts. 
	\item \citep{DBLP:conf/acpr/LinMLLWL17} uses recurrent neural networks with skip-thought vectors to represent news text.
\end{itemize}

Table \ref{tab:index_result} shows the test set results on predicting the S\&P 500 index.
From the table we can see that our ANRES model achieves the best results on the test sets.
By comparing with \citep{luss2015predicting}, we can find that using news event embeddings and deep learning modules can be better representative and also flexible when dealing with high-dimension features.

When comparing with \citep{DBLP:conf/ijcai/DingZLD15} and the knowledge-enhanced \citep{DBLP:conf/coling/DingZLD16}, we find that extracting structured events may suffer from error propagation.
And more importantly, modeling the correlations between trading days can better capture the compositional effects of multiple news events.

By comparing with \citep{DBLP:conf/acl-alta/PinheiroD17} and \citep{DBLP:conf/acpr/LinMLLWL17}, despite that modeling the correlations between trading days can bring better results, we also find that modeling the noise by using a state-related random factor may be effective because of the high market stochasticity.

\begin{table}
\centering
\small
\setlength{\tabcolsep}{1mm}{
\begin{tabular}{l|c|c}
\hline
 & \textbf{Accuracy} & \textbf{MCC} \\
\hline
\citep{luss2015predicting} & 56.38\% & 0.0711 \\
\citep{DBLP:conf/ijcai/DingZLD15} & 64.21\% & 0.4035 \\
\citep{DBLP:conf/coling/DingZLD16} & 66.93\% & 0.5072 \\
\citep{DBLP:conf/acl-alta/PinheiroD17} & 63.34\% & - \\
\citep{DBLP:conf/acpr/LinMLLWL17} & 64.55\% & - \\
\hline
ANRES & \textbf{67.34\%} & \textbf{0.5475} \\
\hline
\end{tabular}
}
\vspace{-0.1cm}
\caption{
\label{tab:index_result}
Test set results on predicting S\&P 500 index.
}
\end{table}

\subsection{Ablation Study on News and Noise}

We explore the effects of different types of news events and the introduced random noise factor with ablation on the test set.
More specifically, we disable the past news, the present news, future news and the noise factor, respectively.
The S\&P 500 index prediction results of the ablated models are shown in Table \ref{tab:ablation_study}.
First, without using the past news events, the result becomes the lowest.
The reason may be that history news contains the biggest amount of news events.
In addition, considering the trading sequence length and the time windows of future news, if we disable the past news, most of them will not be involved in our model at any chance, while the present or the past news will be input on adjacent trading days.

Second, it is worth noticing that using the future news events is more effective than using the present news events.
On the one hand, it confirms the importances to involve the future news in our ANRES model, which can deal with insider trading factors to some extent.
On the other hand, the reason may be the news impact redundancy in sequence, as the future news impact on the $t-1$-th day should be transited to the $t$-th day to compensate the absent loss of the present news events.

The effect of modeling the noise factor is lower only to modeling the past news events, but higher than the other ablated models, which demonstrates the effectiveness of the noise factor module.
We think the reason may because that modeling such an additive noise can separate the effects of news event impacts from other factors, which makes modeling the stock price movement trends more clearly.

\begin{table}
\centering
\small
\begin{tabular}{l|c|c}
\hline
 & \textbf{Acc} & \textbf{MCC} \\
\hline
w/o Past News & 62.17\% & 0.4421 \\
w/o Present News & 64.73\% & 0.4823 \\
w/o Future News & 64.58\% & 0.4781 \\
w/o Noise & 63.90\% & 0.4608 \\
ANRES & \textbf{67.34\%} & \textbf{0.5475} \\
\hline
\end{tabular}
\vspace{-0.1cm}
\caption{
\label{tab:ablation_study}
Test set results of ablation study.
}
\end{table}


\subsection{Predicting Individual Stock Movements}

\begin{table*}
\centering
\small
\setlength{\tabcolsep}{1mm}{
\begin{tabular}{l|l|c|c|c|c|c|c|c|c}
\hline
\multirow{2}{*}{\textbf{Stock}} & \multirow{2}{*}{\textbf{Sector}} & \multicolumn{3}{c|}{\textbf{Company News}} & \multicolumn{3}{c|}{\textbf{Sector News}} & \multicolumn{2}{c}{\textbf{All News}} \\
\cline{3-10}
 & & \textbf{\#docs} & \textbf{Accuracy} & \textbf{MCC} & \textbf{\#docs} & \textbf{Accuracy} & \textbf{MCC} & \textbf{Accuracy} & \textbf{MCC} \\
\hline
Apple & IT & 2,398 & 69.21\% & 0.5632 & 12,812 & 64.35\% & 0.3861 & 56.14\% & 0.2355 \\
Citigroup & Financials & 2,058 & 63.57\% & 0.5193 & 117,659 & 56.29\% & 0.3021 & 55.15\% & 0.1852 \\
Boeing Company & Industrials & 1,870 & 66.25\% & 0.4423 & 17,969 & 61.35\% & 0.2719 & 57.23\% & 0.1824 \\
Google & Communication & 1,762 & 66.13\% & 0.3717 & 13,344 & 60.47\% & 0.2644 & 58.41\% & 0.1387 \\
Wells Fargo & Financials & 845 & 61.64\% & 0.3944 & 117,659 & 57.34\% & 0.1294 & 54.64\% & 0.0823 \\
\hline
\end{tabular}
}
\vspace{-0.1cm}
\caption{
\label{tab:stocks}
Test set results of individual stock price movement prediction.
}
\end{table*}

\begin{figure*}
\centering
\includegraphics[width=15.4cm]{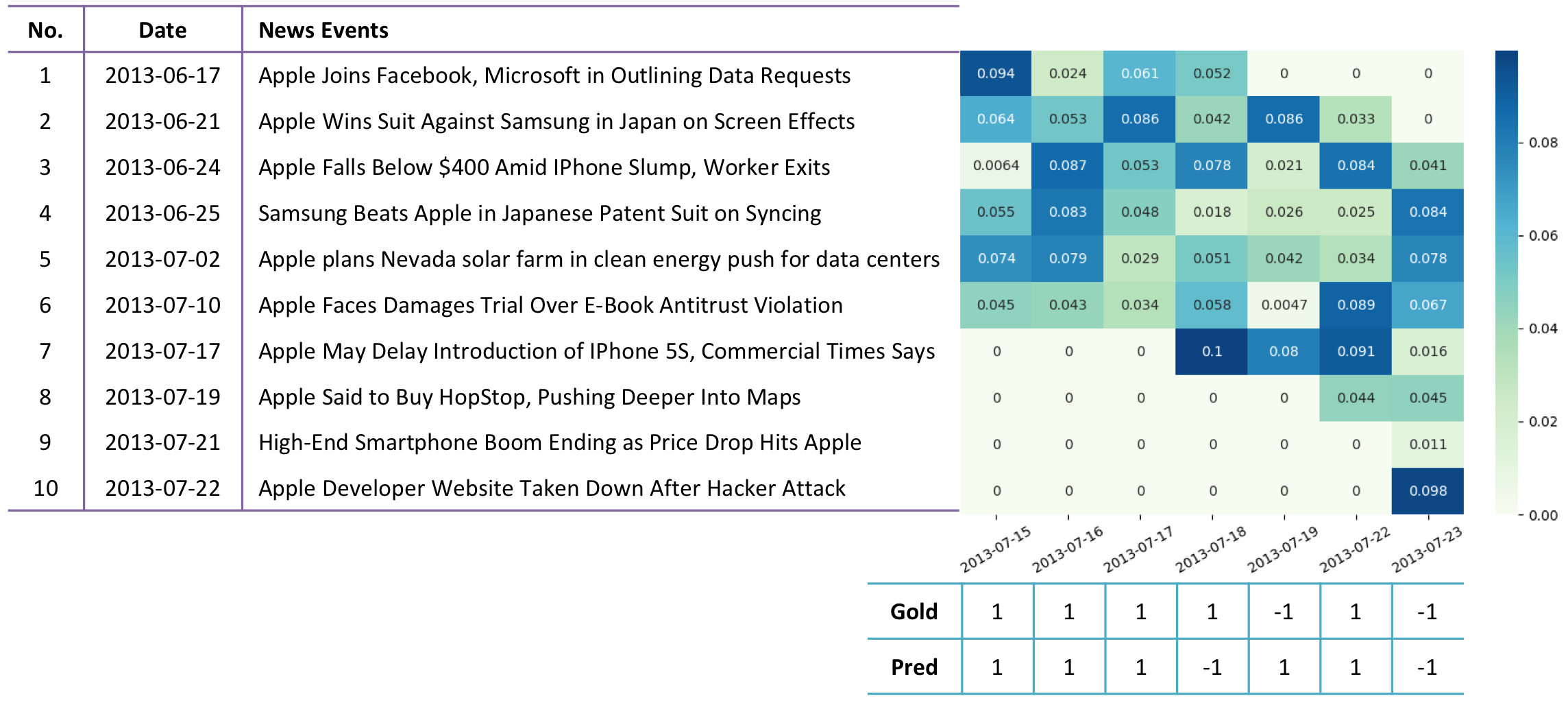}
\vspace{-0.1cm}
\caption{
Attention visualization and test set results comparison of the trading sequence $[$07/15/2013, 07/23/2013$]$ when predicting Apple Inc.'s stock price movements using only company news.
}
\label{fig:interpreting}
\end{figure*}

Other than predicting the S\&P 500 index, we also investigate the effectiveness of our approach on the problem of individual stock prediction using the test set.
We count the amounts of individual company related news events for each company by name matching, and select five well known companies with sufficient news, \textit{Apple}, \textit{Citigroup}, \textit{Boeing Company}, \textit{Google} and \textit{Wells Fargo} from four different sectors, which is classified by the Global Industry Classification Standard.
For each company, we prepare not only news events about itself, but also news events about the whole companies in the sector.
We use company news, sector news and all financial news to predict individual stock price movements, respectively.
The experimental results and news statistics are listed in Table \ref{tab:stocks}.

The result of individual stock prediction by only using company news dramatically outperforms that of sector news and all news, which presents a negative correlation between total used amounts of news events and model performance.
The main reason maybe that company-related news events can more directly affect the volatility of company shares, while sector news and all news contain many irrelevant news events, which would obstruct our ANRES model's learning the underlaying stock price movement trends.

Note that \citep{DBLP:conf/ijcai/DingZLD15,DBLP:conf/coling/DingZLD16} and \citep{DBLP:journals/corr/abs-1902-04994} also reported results on individual stocks.
But we cannot directly compare our results with them because the existing methods used different individual stocks on different data split to report results, and \citep{DBLP:conf/ijcai/DingZLD15,DBLP:conf/coling/DingZLD16} reported only development set results.
This is reasonable since the performance of each model can vary from stock to stock over the S\&P 500 chart and comparison over the whole index is more indicative.


\subsection{Case Study}
To look into what news event contributes the most to our prediction result, we further analyze the test set results of predicting \textit{Apple Inc.}'s stock price movements only using company news, which achieves the best results among the five selected companies mentioned before.

As shown in Figure \ref{fig:interpreting}, we take the example trading sequence from 07/15/2013 to 07/23/2013 for illustration.
The table on the left shows the selected top-ten news events, while attention visualization and results are shown on the right chart.
Note that there are almost fifty different past news events in total for the trading sequence, and the news events listed on the left table are selected by ranking attention scores from the past news events, which are the most effective news according to the ablation study.
There are some zeros in the attention heat map because these news do not belong to the corresponding trading days.

We can find that the news event No. 1 has been correlated with the stock price rises on 07/15/2013, but for the next two trading days, its impact fades out.
On 07/18/2013, the news event No. 7 begins to show its impact.
However, our ANRES model pays too much attention in it and makes the incorrect prediction that the stock price decreases.
On the next trading day, our model infers that the impact of the news event No. 2 is bigger than that of the news event No. 7, which makes an incorrect prediction again.
From these findings, we can see that our ANRES model tends to pay more attention to a new event when it first occurs, which offers us a potential improving direction in the future.

\section{Conclusion}
We investigated explicit modeling of stock value sequences in news-driven stock prediction by suing an LSTM state to model the fundamentals, adding news impact and noise impact by using attention and noise sampling, respectively.
Results show that our method is highly effective, giving the best performance on a standard benchmark.
To our knowledge, we are the first to explicitly model both events and noise over a fundamental stock value state for news-driven stock movement prediction.

%

\bibliographystyle{plain}
\bibliography{stock}

\end{document}